\documentclass{IOS-Book-Article}

\usepackage{soul}\setuldepth{article}
\usepackage{mathptmx}
 \usepackage{graphicx}
 \usepackage{pgfplots}
\usepackage{multirow}
\usepackage{adjustbox}

\usepackage{caption} 
\def\hb{\hbox to 10.7 cm{}}

\begin{document}

\pagestyle{headings}
\def\thepage{}

\begin{frontmatter}              

\title{Analysing the Performance of Stress Detection Models on Consumer-Grade Wearable Devices}

\markboth{}{April 2021\hb}
%
\author[A]{\fnms{Van-Tu} \snm{Ninh}},
\author[B]{\fnms{Sinéad} \snm{Smyth}},
\author[C]{\fnms{Minh-Triet} \snm{Tran}},
and
\author[A]{\fnms{Cathal} \snm{Gurrin}}


\runningauthor{Van-Tu Ninh et al.}
\address[A]{School of Computing, Dublin City University}
\address[B]{School of Psychology, Dublin City University}
\address[C]{VNU-HCM, University of Science}

\begin{abstract}
 Identifying stress level can provide valuable data for mental health analytics as well as labels for annotation systems. 
 Although much research has been conducted into stress detection models using heart rate variability at a higher cost of data collection, there is a lack of research on the potential of using low-resolution Electrodermal Activity (EDA) signals from consumer-grade wearable devices to identify stress patterns. 
 In this paper, we concentrate on performing statistical analyses on the stress detection capability of two popular approaches of training stress detection models with stress-related biometric signals: user-dependent and user-independent models.
 Our research manages to show that user-dependent models are statistically more accurate for stress detection.  
 In terms of effectiveness assessment, the balanced accuracy (BA) metric is employed to evaluate the capability of distinguishing stress and non-stress conditions of the models trained on either low-resolution or high-resolution Electrodermal Activity (EDA) signals. 
 The results from the experiment show that training the model with (comparatively low-cost) low-resolution EDA signal does not affect the stress detection accuracy of the model significantly compared to using a high-resolution EDA signal.
 Our research results demonstrate the potential of attaching the user-dependent stress detection model trained on personal low-resolution EDA signal recorded 
 to collect data in daily life to provide users with personal stress level insight and analysis.
\end{abstract}

\end{frontmatter}
\markboth{January 2020\hb}{January 2020\hb}

\section{Introduction}
Recently, there has been a significant volume of research into personal sensing, with various applications in self-quantification, lifelogging, and healthcare \cite{Gurrin2014LifeLoggingPB}. 
Specifically, as wearable devices for health tracking via physiological data are becoming more popular, researchers are beginning to exploit such kinds of personal data and analyze them to provide not only insights into an individual's daily life activities and movement \cite{Gurrin2016OverviewON}, but also meaningful insights into both physical and mental health status.
For example, accelerometer and gyroscope data generated from wearable devices and smartphones can be used to recognize human activities \cite{Le2020SystematicEO, Bhat2018OnlineHA}.
In addition, heart rate (HR) and electrodermal activity (EDA)/skin response (SR) are discriminative signals that can provide clues to personal affective states and stress levels  \cite{Can2019ContinuousSD}. 
Many approaches of building an automatic emotion state/stress level discriminator using biometric data have been proposed, but most of them investigate the performance of the detection model using professional grade, high-resolution devices in controlled laboratory settings \cite{Wang2017AccuracyOW, Schmidt2018IntroducingWA, Nkurikiyeyezu2019TheIO}. 
Moreover, although the performance of stress detection models using Heart Rate (HR) and Heart Rate Variability (HRV) from wearable devices as well as data validation of HR signal and HRV are approved by many works \cite{Menghini2019StressingTA, Schuurmans2020ValidityOT, Lier2019ASV, Wang2017AccuracyOW, Milstein/fnbeh.2020.00148, Kim2018StressAH}, there is limited use of low-resolution (consumer-grade) EDA signals recorded from wearable devices and little is known on the resultant effect on the performance of stress detection \cite{PPR:PPR233643, Nkurikiyeyezu2019TheIO, Schmidt2018IntroducingWA, Siirtola2020ComparisonOR, Healey2005DetectingSD, Zangroniz2017CD}. Therefore, in this paper, we concentrate on studying and comparing different approaches of constructing stress detection models using low-resolution EDA signals.


Stress is defined as a physical, mental, or emotional response of the body to anything that requires attention or action \cite{Schmidt2018IntroducingWA} and can be categorized into three main types depending on the level of impact it can have: acute stress, episodic acute stress, and chronic stress. 
As chronic stress and anxiety are risk factors for dementia and cardiovascular diseases during aging \cite{Gimson2018SupportFM}, there is a need for some form of acute stress detection system that can monitor the stress level of an individual in daily life, either to generate a source of metadata for personal health data, or as a trigger for therapies to alleviate its harmful effects.  
Therefore, more work on daily-life stress detection using ubiquitous low-cost devices needs to be carried out to explore promising techniques and challenges in this area.


There are two common challenges when building a stress detection model. 
One of the challenges is that the stress monitoring system needs to adapt to the physiological data of every individual since people have different physiological reactions to stress according to  \cite{Schmidt2018WearableAA}, therefore, it is likely that stress detection models operate more accurately when it is customised for each individual, rather than for a population.
However, no work has been done to provide a conclusion that which of these two stress detection models is an optimal 
in terms of proving the conclusion does not happens by random chance.
Especially, no comparison and conclusion have been made to determine whether a user-dependent or user-independent approach is better when training stress detection models on low-resolution signals.  
Accordingly, there is a need for a comparison between the effect of different signal-resolution on the model's stress level discrimination capability. 

In summary, we present three main contributions of this paper:
\begin{itemize}
    \item From a sample of participants' data, we show that using low-resolution EDA signal as input for a stress detection model maintains the same performance of the models as high-resolution EDA signals in the same condition of the study protocol, regardless of the choice of learning models.
    \item With supportive evidence, we show that stress detection models with low-resolution EDA signal input from both wrist-worn device and finger-mounted sensor manage to achieve high balanced accuracy (BA) scores ranging from $66.10\%$ to $100\%$ with mean and median BA of $90.53\%$ and $93.00\%$ respectively using Support Vector Machine models. This also shows the potential of using EDA data recorded from sensors of wearable devices to keep track of stress level.
    \item Additionally we show that user-dependent stress detection models are statistically more accurate than user-independent ones, regardless of the choice of learning models.
\end{itemize}

\section{Related Work} \label{st:RelatedWork}
Heart rate (HR) and Electrodermal Activity (EDA) were found to be discriminative signals for stress level measurement \cite{Can2019ContinuousSD}. 
Although many data validation works were performed and approved the efficiency of using HR and HRV of wearable devices in stress detection problems, there was not much research on the efficiency of EDA signal of wearable devices in stress detection models.
Specifically, Luca Menghini et al. showed that there is neither correlation nor visual resemblance between wrist-EDA and finger-EDA measurement and suggested more studies should be performed to assess the responsiveness of wrist-EDA signal to emotional and cognitive stress \cite{Menghini2019StressingTA}. 
Nir Milstein et al. concluded that the low-resolution EDA data recorded from Empatica E4 wrist-band, which is a popular wrist-worn device for real-time physiological data streaming, is not reliable \cite{Milstein/fnbeh.2020.00148}. 
These are the motivations for us to conduct more experiments to validate the effect of low-resolution EDA signal recorded by wearable devices on the stress detection accuracy of the models.


In 2018, Philip Schmidt et al. introduced a new publicly available dataset named WESAD for wearable stress detection and provided  preliminary work on their new dataset \cite{Schmidt2018IntroducingWA}.
According to their work, the Linear Discriminant Analysis (LDA) model achieved the highest accuracy score of 93.12\% for binary (stress/non-stress) classification using multimodal physiological signals recorded by a chest-worn device. 
However, their evaluation metric using accuracy was not appropriate due to the unequal number of stress and non-stress labels in the WESAD dataset which induced bias when evaluating the results. 
Pekka Siirtola continued this work by analysing the performance of user-independent stress detection models and the effect of window-size on recognition rates using low-resolution signals recorded by Empatica E4 wrist-band only \cite{Siirtola2019ContinuousSD}. 
Their best balanced accuracy score of 87.4\% was achieved by training LDA model with the combination of Skin Temperature (ST), Blood Volume Pulse (BVP), and HR. 
However, the best balanced accuracy score of the Random Forest model trained on EDA signal was only $78.3\%$. 

Kizito Nkurikiyeyezu et al. conducted experiments on the performance of these two models using chest-worn HRV and EDA signals from both WESAD \cite{Schmidt2018IntroducingWA} and SWELL datasets \cite{SWELLKoldijk2014} trained with a well-defined-hyperparameter Random Forest model \cite{Nkurikiyeyezu2019TheIO} to compare user-dependent and user-independent models. 
To avoid the pitfall when choosing accuracy as an evaluation metric, they down-sampled the dataset by randomly discarding samples of majority classes to balance the number of categories. 
Through their experiment, the user-dependent models achieved higher accuracy than user-independent ones in both datasets. 
They also proposed a hybrid calibrated model to improve the performance of the user-independent model from $42.5\% \pm 19.9\%$ to $95.2\% \pm 0.5\%$ by including a small number of samples of the unseen subject $(n = 100)$ \cite{Nkurikiyeyezu2019TheIO}. Additionally, Pekka Siirtola et al. also made a comparison of user-independent and user-dependent stress detection models using the AffectiveROAD \cite{HaouijAffectiveRoad2018} dataset \cite{Siirtola2020ComparisonOR}. 
They suggested applying a subject-wise feature selection to improve user-independent model instead of purely building personal models using personal training data. 
The average balanced accuracy scores of both user-independent and user-dependent stress detection models were not encouraging, especially when using low-resolution EDA data only.
Therefore, in our work, we propose an optimal approach of constructing high-accuracy stress detection models including feature extraction process with detailed description of the model training methods by proving the two hypotheses mentioned in section \ref{sst:InitialHypotheses}.  


\section{Method}
\subsection{Initial Hypotheses} \label{sst:InitialHypotheses}
Our experiments are conducted to validate the impact of EDA signal's quality recorded from both wrist-worn device and finger-mounted sensors on the prediction capability of stress detection models to prove that the proposed method for the stress detection model construction is trustworthy.
Moreover, we evaluate the stress detection balanced accuracy of user-independent and user-dependent models statistically to propose an optimal approach of training stress detection models with low-resolution EDA signal. These two tasks are main contributions of our work, which are completed by either proving or disproving the two following hypotheses:
\begin{itemize}
    \item \textit{Hypothesis 1}: Using low-resolution EDA signal, the user-dependent stress detection model is statistically more accurate than the user-independent one in terms of discriminating stress an non-stress patterns.
    \item \textit{Hypothesis 2}: The stress detection accuracy of the user-dependent model trained on low-resolution EDA signal is not statistically different from the one trained on high-resolution EDA signal, which implies that the EDA signal from consumer-grade wearable devices does not deteriorate stress detection accuracy of the model.
\end{itemize}

\subsection{Datasets} \label{st:Datasets}
In our experiments, we use the WESAD \cite{Schmidt2018IntroducingWA}, AffectiveROAD \cite{HaouijAffectiveRoad2018}, and DCU-NVT-EXP1 datasets to prove hypothesis 1 and 2.
All three datasets are used to compare the stress detection accuracy of user-independent and user-dependent models statistically, which addresses the first hypothesis.
The second hypothesis is addressed using WESAD with supportive results from AffectiveROAD and DCU-NVT-EXP1 datasets as only WESAD dataset contains the records of synchronized high-sampling rate (700 Hz) and low-sampling rate data (4 Hz).

\textbf{WESAD dataset}: The released dataset consisted of physiological data collected from 15 participants under two different study protocols including a combination of amusement/stress/relaxation conditions. 
Philip Schmidt et al. recorded high-resolution physiological Blood Volume Pulse (BVP), Electrocardiogram (ECG), Electrodermal activity (EDA), Electromyogram (EMG), Respiration (RESP), Skin temperature (TEMP), and motion (ACC) modalities from a chest-worn device named RespiBAN. 
Concurrently, participants were also required to wear the Empatica E4 wrist-band to capture low sampling-rate heart rate and EDA data simultaneously. 
The authors used study protocol as ground-truth. 
In our experiment, we consider the chest-worn data as WESAD-Chest and the wrist-worn data as WESAD-Wrist. 

\textbf{AffectiveROAD dataset}: The dataset was collected to identify drivers' state indicators such as stress and arousal. 
Neska El-Haouij et al. gathered low-resolution BVP and EDA data from an Empatica E4 wrist-band as well as high-resolution ECG and RESP data using a chest-worn device called Zephyr BioHarness 3.0. 
10 participants were invited to join in 14 different driving tasks. 
The stress level of the participant was rated on a continuous "stress" metric ranging from 0 (no stressful) to 1 (extremely stressful) by the experimenter using a slider sitting in the rear of the car during each driving task. 
The subjective stress level was then validated again by the participants. 
In our work, we choose the optimal threshold of $0.4$ to divide the continuous "stress" metric into stress/non-stress labels as Pekka Siirtola et al. suggested in their work. 
This is used as ground-truth of AffectiveROAD dataset in our experiment.


\textbf{DCU-NVT-EXP1 dataset}: To increase the number of samples to prove hypothesis 1 and support hypothesis 2 that user-dependent models trained with low-resolution EDA signal can provide competitive stress prediction accuracy, 
we created a dataset named DCU-NVT-EXP1 using finger-mounted sensor MINDFIELD eSense Skin Response (5 Hz) to record low-resolution EDA signal of participants during a study protocol containing both daily-life tasks and Virtual Reality tasks in a pilot study. 
The first stress test was the Trier Social Stress Test (TSST)  \cite{Kirschbaum1993TheS}.
The second stress test was the Sing-a-Song Stress Test (SSST) \cite{VANDERMEE2020102612} and the final one was the Virtual Reality (VR) Stress Test including five stress-stimulated VR games.
In total, 7 participants, which are post-graduate students in Dublin City University with no-experience on stress-related experiments, 
joined the study protocol and conducted self-evaluation forms including Subjective Units of Distress Scale (SUDS) for subjective stress evaluation and NASA Task Load Index (NASA-TLX) for an overall workload score.
The subjective stress self-evaluation was then divided into four degrees of stress which were relaxed (0-20), mildly-stressed (30-50), stressed (60-80), and extremely stressed (90-100).
From this division, we continued to categorize them into binary categories including non-stress (relaxed \& mildly-stressed) and stress (stress \& extreme stress) with the assumption that the target-user should be warned about his/her stress status when he/she experiences high degree of anxiety or distress. 
The binary categories of stress/non-stress status were used as the ground-truth in this dataset. 


\subsection{Statistical EDA Feature Extraction} \label{sst:EDAFeatureExtraction}
The EDA signal was filtered through a 5 Hz fourth-order Butterworth low-pass filter if its Nyquist frequency is in range of $0$ and $1$. In the context of our experiment, this means that only the high-resolution EDA signal from WESAD-Chest data is pre-processed through a low-pass filter before extracting the feature. 
Then, EDA features which comprise of Skin Conduction Response (SCR), Skin Conduction Level (SCL), SCR Peaks, SCR Onsets, and SCR Amplitude were extracted. 
Finally, the statistical features from four works in the same field \cite{Choi2012DevelopmentAE, Nkurikiyeyezu2019TheIO, Schmidt2018IntroducingWA, Healey2005DetectingSD} were computed except for the slope of EDA signal along the time-axis. 
The window-size and window-shift used in this feature extraction process were 60 seconds and 30 seconds respectively, which means that the model searches for stress pattern after each 30 seconds by observing the statistical features extracted from signals in the 60-second interval from the current signal point.  


\subsection{Stress Detection Model Training} \label{sst:StressDetectionModelsTraning}
Using the aforementioned statistical features extracted in section \ref{sst:EDAFeatureExtraction}, we train stress detection models using Grid Search on five regularly used Machine Learning models which are Logistic Regression (LR), Random Forest (RF), Support Vector Machine (SVM), Multi-layer Perceptron (MLP), and K-Nearest Neighbors (KNN).
Standardization of the feature along its axis is applied for the input of SVM, MLP, and KNN as these Machine Learning models work effectively on standardized data. 
We perform Leave-One-Group-Out (LOGO) strategy for the user-independent stress detection model. 
The training and test data for user-dependent training is divided using a stratified train-test split strategy whose test size equals to 28.6\%. 
We are also aware of the imbalanced characteristic of the datasets when training the models by setting \textit{class\_weight}
parameters in Grid Search. 
The Grid Search parameters for each Machine Learning model are shown in Table \ref{tab:GridSearchConfiguration}. 
\begin{table}[h!]
    \centering
    \caption{Grid Search configurations of five Machine Learning models used when training stress detection models. 
    \label{tab:GridSearchConfiguration}}
    \begin{tabular}{ | c | c | c |} 
      \hline
      \textbf{Model} & \textbf{GridCV Parameters} & \textbf{Values} \\
      \hline
      \multirow{2}{2em}{LR} & C (Regularization) & 0.001, 0.01, 0.1, 0.25, 0.5, 0.75, 1, 10 \\
      & class\_weight & None, balance \\
      \hline
      \multirow{5}{2em}{RF} & n\_estimators & 500, 1000 \\
      & min\_samples\_split & 2, 4 \\
      & min\_samples\_leaf & 1, 4 \\
      & class\_weight & None, balance \\
      \hline
      \multirow{2}{2em}{SVM} & C (Regularization) & 0.001, 0.01, 0.1, 0.25, 0.5, 0.75, 1, 10 \\
      & class\_weight & None, balance \\
      \hline
      \multirow{1}{2em}{MLP} & hidden\_layer\_sizes & 64, 128, 256, 512 \\
      \hline
      \multirow{2}{2em}{KNN} & n\_neighbors & 3, 5, 7 \\
      & weights & uniform, distance \\
      \hline
    \end{tabular}
\end{table}

\subsection{Statistical Analysis}
Although many works show the accuracy and precision of stress detection models using low-resolution EDA signal data \cite{Nkurikiyeyezu2019TheIO, Schmidt2018IntroducingWA, Zangroniz2017CD, Siirtola2020ComparisonOR}, they only provide accuracy and precision score of detection models on specific datasets without performing statistical analyses on the results. 
In our experiment, we concentrate on providing conclusions based on inferential statistics, which draw conclusions about the population based on a number of representative samples. 
Specifically, hypothesis testing is used to estimate statistical performance of stress detection models using low-resolution EDA data recorded by available commercial wrist-worn devices or finger-mounted sensors compared to chest-worn/clinical ones based on three representative datasets as well as comparing the statistical stress detection capability between user-independent and user-dependent models.  
    
\section{Experimental Results}
\subsection{Evaluation Metric}
As most of these datasets have unequal numbers of stress/non-stress labels, a proper evaluation metric is chosen to avoid bias assessment of experimental results. 
Although many pre-processing techniques can be applied to transform an imbalanced dataset into a balanced one, such as Synthetic Minority Over-sampling Technique (SMOTE) and Adaptive Synthetic Sampling Approach (ADASYNC), we do not apply these techniques to balance the dataset by removing features of dominant category since important information can be discarded through this process. 
Based on the analysis of Sirko Straube et al., balanced accuracy (BA) is an appropriate choice to evaluate results as we prefer assessing the capability of distinguishing the two categories to evaluating the precision of detecting stress patterns only \cite{Straube2014EvaluationMetric}. 
Moreover, it is more intuitive to use one evaluation metric in statistical analysis than using a combination of different evaluation scores such as accuracy, precision, recall, etc. to assess the statistical performance of a detection/prediction model with respect to one of its components or the attribute of inputs.

\subsection{Statistical Analysis on the Stress Detection Capability of User-Dependent and User-Independent Models} \label{sst:ComparisonUDvsUI}
Although the experiment made by Kizito Nkurikiyeyezu et al. concluded that the user-dependent stress detection model manages to predict more accurately than the user-independent one, they only tested it on high-resolution data in the SWELL \cite{SWELLKoldijk2014} and WESAD-Chest datasets \cite{Schmidt2018IntroducingWA} and trained them using a simple Random Forest model \cite{Nkurikiyeyezu2019TheIO}.
Therefore, we conduct experiments on the three datasets (WESAD-Wrist, DCU-NVT-EXP1, and AffectiveROAD) which have low-resolution EDA signal.
As mentioned in section \ref{sst:InitialHypotheses}, we expect that user-dependent model detects stress patterns more accurately than user-independent one statistically, regardless of the choice of Machine Learning models. 

Five user-independent and five user-dependent Machine Learning models are trained using a LOGO strategy with GridCV configurations, as described in section \ref{sst:StressDetectionModelsTraning}.
Balanced accuracy scores are then computed for each individual in each dataset to create evaluation scores of both user-independent and user-dependent models respectively. 
Finally, hypothesis testing is conducted to either prove or disprove the initial hypothesis and provide the estimation of the difference between the population of user-dependent and user-independent Machine Learning models' performance trained on low-resolution EDA signals based on a random sample of data. 
The data contains $2 \times 5 \times 35 = 350$ observations in total for both outputs from user-dependent and user-independent models. 
Each consists of independent outputs of $5$ Machine Learning models $\times 35$ participants’ data in three datasets: AffectiveROAD, WESAD-Wrist, and DCU-NVT-EXP1.

\begin{figure}[h]
    \centering
    \includegraphics[width=0.8\linewidth]{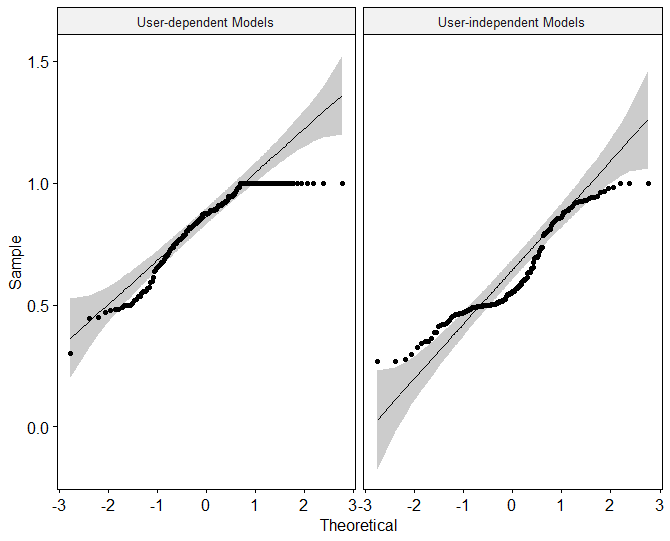} 
    \caption{QQ-Plot of the performance of user-dependent and user-independent stress detection models trained on low-resolution EDA signal}
    \label{fig:QQPlotRQ1-1}
\end{figure}

Firstly, we employ Shapiro-Wilk test (SW) to test the normality of two independent data from two groups as it is proved to be the most powerful test for data normality \cite{Yapnormality}. 
Moreover, the Anderson-Darling test is also utilized to support the conclusion of the normality of this data. 
In these tests, the distribution of non-absolute difference data (BA scores of user-dependent model minus BA scores of user-independent model) is compared with the normal distribution where null hypothesis $H_0$ assumes that the data comes from  the normal distribution.
The p-values of the Shapiro-Wilk test of user-dependent and user-independent group are $2.66 \times 10^{-8}$ and $1.425 \times 10^{-15}$ respectively, which are both smaller than the alpha level of $0.05$. 
In addition, the test statistics of Anderson-Darling test for both user-dependent and user-independent groups are  $6.1268$ and $6.3181$ correspondingly, which are both larger than the critical value of $0.722$ at significant level of $0.05$. 
These values indicate that the distributions of the two sample data are not normal as the null hypothesis of both tests are rejected. 
This can be seen visually in Figure \ref{fig:QQPlotRQ1-1}. 
Therefore, the Wilcoxon rank sum test is applied to prove that the user-dependent model discriminates stress and non-stress patterns statistically better than the user-independent ones. 
This implies that the difference of the medians between the user-dependent and user-independent models should be large and the median of the user-dependent model should be almost always to the right of the one of user-independent model. 
From this point, the null and alternative hypotheses of the one-sided Wilcoxon rank sum test are established as follows:

\begin{itemize}
    \item $H_0: M_1 = M_2$ (The user-dependent model has no improvement in distinguishing stress and non-stress patterns).
    \item $H_a: M_1 > M_2$ (The user-dependent model manages to distinguish stress/non-stress patterns statistically more accurately than user-independent one).
\end{itemize}

\begin{figure}[h]
    \centering
    \includegraphics[width=0.8\linewidth]{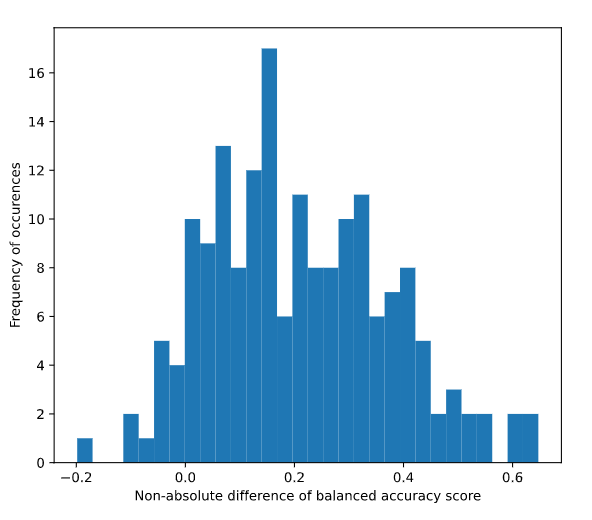} 
    \caption{Distribution of non-absolute difference of balanced-accuracy scores between user-dependent and user-independent model from AffectiveROAD, WESAD-Wrist, and DCU-NVT-EXP1 datasets}
    \label{fig:RQ1Dist}
\end{figure}

In the above hypotheses, $M_1$ and $M_2$ indicate the median of stress/non-stress prediction accuracy scores of user-dependent and user-independent models respectively. 
The p-value of the Wilcoxon rank sum test computed from the test-statistic value of $6303$ is $7.89 \times 10^{-22}$, which is significantly smaller than the pre-determined value of significance of $0.001$. 
This implies that there is enough evidence to reject the null hypothesis with confidence level of $99.9\%$.
This also means that statistically, the user-dependent model outperforms the user-independent one. 
The confidence interval ranges from $-\infty$ to $-0.18$.
This indicates that the median of stress/non-stress prediction accuracy scores of the user-independent model is statistically less than the one of user-dependent model at most $0.18$ with the confidence level of $99\%$.
In addition, the median of the difference between two samples is estimated to be $-0.2401$, which is significant. 
From these findings, the user-dependent model can be concluded to discriminate stress and non-stress patterns statistically more accurately than the user-independent model in terms of balanced accuracy. 
The conclusion can be seen via Figure \ref{fig:RQ1Dist} using histogram plot.

\subsection{Statistical Analysis on the Effect of Signal-resolution of EDA Signal on the Performance of Stress Detection Models} \label{sst:ComparisonLowVsHighEDA}
\begin{figure}[h]
    \centering
    \includegraphics[width=0.8\linewidth]{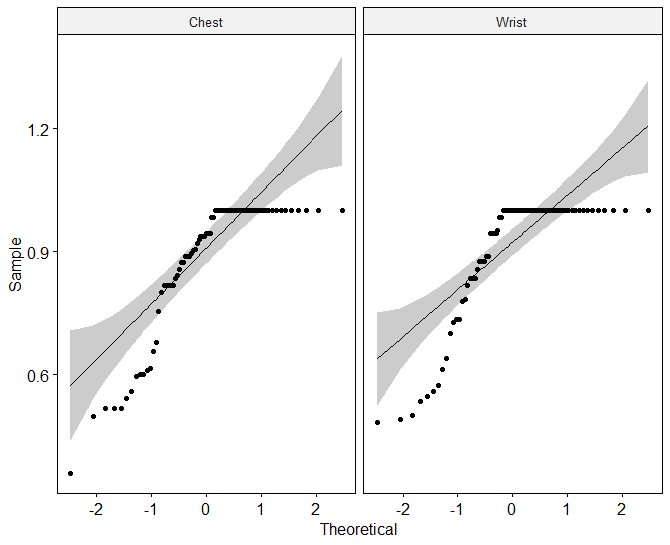} 
    \caption{QQ-Plot of the performance of user-dependent stress detection models trained on WESAD-Chest and WESAD-Wrist datasets}
    \label{fig:QQPlotRQ1-2}
\end{figure}

As the user-dependent stress detection model is proven to be more efficient than user-independent one, therefore, we only consider the evaluation scores of user-dependent model in this experiment. 
As WESAD dataset contains both low-resolution and high-resolution EDA signal, only this data is used in this experiment to compare the stress detection accuracy of the user-dependent model in terms of discriminating stress and non-stress patterns trained on either low-resolution EDA signal from wrist-worn device or high-resolution one from chest-worn/clinical device regardless of different Machine Learning models. 
The approach of testing the hypothesis is the same as in section \ref{sst:ComparisonUDvsUI}. 
Since the normality tests (WESAD-Wrist: Shapiro-Wilk's p-value of $1.09 \times 10^{-9} < 0.05$ and Anderson-Darling's test statistic of $9.985 > 0.722$, WESAD-Chest: Shapiro-Wilk's p-value of $2.04 \times 10^{-11} < 0.05$ and Anderson-Darling's test statistics of $7.003 > 0.722$) indicate that the samples from both independent groups do not follow a normal distribution with the confidence of $95\%$, Wilcoxon rank sum test is applied in this experiment. 
This can be seen visually through the QQ-plot in Figure \ref{fig:QQPlotRQ1-2}. 
The total number of independent observations used in this research question is $2 \times 5 \times 15 = 150$, which includes $5$ Machine Learning models $\times$ $15$ participants $= 75$ observations. 
The hypotheses of the two-sided Wilcoxon rank sum test are stated as follows:

\begin{itemize}
    \item $H_0$: $M_1 = M_2$ (There is no statistical difference between the stress detection accuracy of the models trained on low-resolution EDA signal and the ones trained on high-resolution EDA signal).
    \item $H_a$: $M_1 \neq M_2$ (The difference between the stress detection accuracy of these models is significant).
\end{itemize}

In the above hypotheses, $M_1$ and $M_2$ are the median of the distribution of two independent balanced accuracy scores from five machine learning models trained on WESAD-Chest and WESAD-Wrist respectively. 
$M_1 = M_2$ indicates that the distributions of the two data are the same, and hence have the same median. 
Naturally, the meaning of "the same" is estimated approximately in statistics, which infers that the difference between the two distributions via the difference of the medians is not statistically significant.   

\begin{figure}[h]
    \centering
    \includegraphics[width=0.8\linewidth]{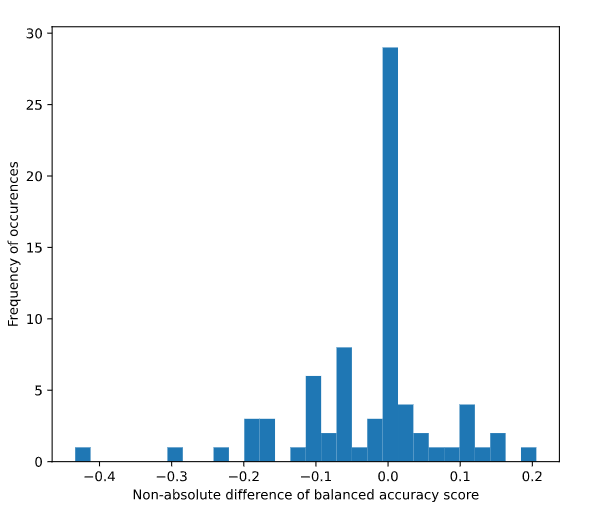} 
    \caption{Distribution of non-absolute difference of balanced-accuracy scores between high-resolution and low-resolution EDA data of WESAD dataset using user-dependent stress detector.}
    \label{fig:RQ1-2Dist}
\end{figure}

Applying the Wilcoxon rank sum test, we obtain the p-value of $0.1491$, which is larger than the pre-determined value of significance of $0.05$. 
The effect size of the test is $0.118$, which is small in terms of magnitude. 
All of the information implies that there is not enough evidence to reject the null hypothesis with confidence level of $95\%$. 
This indicates that there is no significant statistical difference in stress detection accuracy scores when the model is trained on the features of low-resolution EDA signal. 
The $95\%$ confidence interval ranges from $-1.6685 \times 10^{-2}$ to $6.1897 \times 10^{-5}$. 
This indicates that the difference between the median of the accuracy scores of the models using chest-worn EDA sample and the one using wrist-worn EDA data is in that range with the confidence level of $95\%$. 
In addition, the median of the difference between the two samples is estimated to be $-2.7442 \times 10^{-5}$, which is also insignificant. 
Figure \ref{fig:RQ1-2Dist} using histogram plot also implies the conclusion visually. 

\subsection{Accuracy of the User-dependent Stress Detection Model trained on Low-resolution EDA Signal}
\begin{figure}[t]
    \centering
    \includegraphics[width=0.8\linewidth]{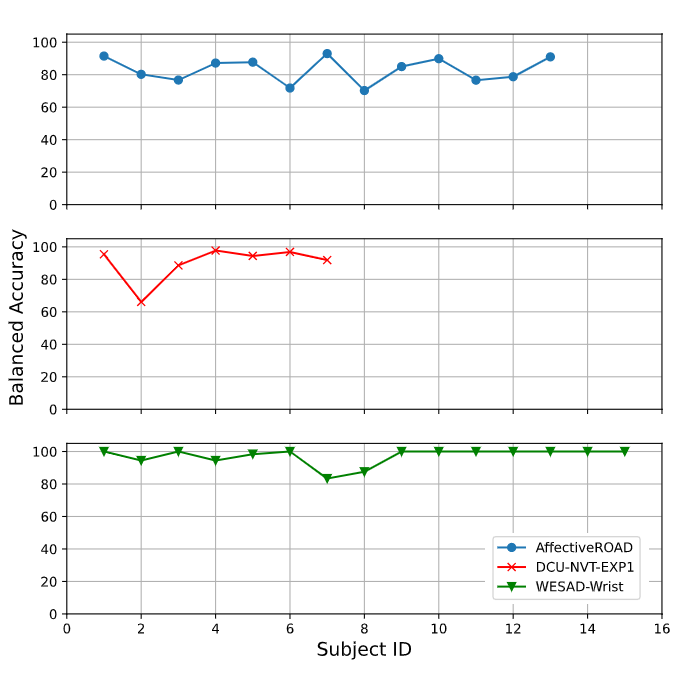} 
    \caption{Balanced accuracy scores of the user-dependent stress detection model built with Support Vector Machine model for each low-resolution EDA signal dataset.}
    \label{fig:BalancedAccuracy}
\end{figure}

Figure \ref{fig:BalancedAccuracy} illustrates supportive evidence for the conclusion in section \ref{sst:ComparisonLowVsHighEDA}. 
As can be seen in figure \ref{fig:BalancedAccuracy}, the balanced accuracy scores of subject-ids in WESAD-Wrist dataset are mostly above 94\%. 
Although two subject data trained with SVM model achieve the balanced accuracy score of $83.33\%$ and $87.5\%$, these scores are also good enough for a stress detection model to be used in commercial product. 
On the AffectiveROAD dataset, the evaluation scores of the model range from $70.24\%$ to $93.00\%$ with its mean of $81.13\%$, which is lower than the results of WESAD-Wrist dataset although the same kind of device is used to record the  EDA signal. 
The balanced accuracy scores of subjects in DCU-NVT-EXP1 dataset are all higher than $88\%$ with the best accuracy score that the model can achieve up to $97.73\%$, except for subject id 2. 
In summary, the mean and the median accuracy score of the user-dependent stress detection model among the three datasets using low-resolution EDA signal trained with SVM model are approximately $90.53\%$ and $93.00\%$ respectively, which show the potential of integrating stress detection into everyday personal data analytics so that more health insight can be obtained. 

\section{Conclusion}
In this paper, we compare the performance of different stress detection models using EDA signal data recorded by both wrist-worn/finger-mounted devices and chest-worn device and apply statistical analysis to conclude the effect of EDA signal quality on the performance of stress detection models statistically. 
Through our experiment, we manage to show that with low-resolution EDA signal data, a user-dependent stress detection model statistically provides more accurate stress detection results than user-independent one.
In addition, we also prove that the stress detection accuracy scores of the model trained on low-resolution EDA signal has no statistically significant difference from the ones of the model trained on high-resolution one recorded by chest-worn/clinical data. 
The conclusion is supported by the balanced accuracy scores of stress detection SVM model trained on low-resolution EDA signal ranging from $66.10\%$ to $100\%$ with mean and median BA of $90.53\%$ and $93.00\%$ respectively.
This also means that the EDA low-resolution signal from consumer-grade devices is good enough to build user-dependent stress detection models, which makes it easier to collect data in daily life for mental health tracking and analysis.
In future work, we aim to combine multi-modal data from lifelog recording to provide context of the stress stimuli to gain insight of the cause so that proper mental health tracking and stress alleviation can be proposed to improve personal mental health in daily life. 
\section{Acknowledgements}
This research was conducted with the financial support of ADAPT Core under Grant Agreement No. 13/RC/2106 at the ADAPT SFI Research Centre at Dublin City University.  The ADAPT SFI Centre for Digital Content Technology is funded by Science Foundation Ireland through the SFI Research Centres Programme and is co-funded under the European Regional Development Fund (ERDF) through Grant Number 13/RC/2106\_P2.
\bibliographystyle{vancouver}
\bibliography{ios-book-article.bib}
\end{document}